\def\year{2020}\relax
\documentclass[letterpaper]{article} 
\usepackage{aaai20}  
\usepackage{times}  
\usepackage{helvet} 
\usepackage{courier}  
\usepackage[hyphens]{url}  
\usepackage{graphicx} 
\usepackage{graphicx}  
\makeatletter
\newcommand{\printfnsymbol}[1]{%
  \textsuperscript{\@fnsymbol{#1}}%
}
\makeatother

 \newcommand{\bb}{{\bf{b}}}
 \newcommand{\bx}{{\bf{x}}}
 \newcommand{\by}{{\bf{y}}}
 \newcommand{\bu}{{\bf{u}}}
 
 \newcommand{\bv}{{\bf{v}}}
 \newcommand{\bW}{{\bf{W}}}
 \newcommand{\R}{{\mathbb{R}}}

\usepackage{xcolor}
\usepackage{subcaption}
\usepackage{amsmath, cuted}
\usepackage{cleveref}
\usepackage{amssymb}
\usepackage{algorithmic}
\usepackage{adjustbox}
\usepackage{booktabs}
 \usepackage{makecell}
\usepackage[para,online,flushleft]{threeparttable}

\title{Dropout Prediction over Weeks in MOOCs \\ by Learning Representations of Clicks and Videos}
\newcommand\equalcontrib{\thanks{Equal contribution}}
\author{
\Large \textbf{Byungsoo Jeon\textsuperscript{\rm 1}\equalcontrib, Namyong Park\textsuperscript{\rm 1}\printfnsymbol{1}}\\ 
\textsuperscript{\rm 1}School of Computer Science, Carnegie Mellon University, Pittsburgh, PA\\
\textsuperscript{\rm 1}\{byungsoj,namyongp\}@cs.cmu.edu\\
}
 
\begin{document}

\maketitle

\begin{abstract}

This paper addresses a key challenge in MOOC dropout prediction, namely to build meaningful representations from clickstream data. While a variety of feature extraction techniques have been explored extensively for such purposes, to our knowledge, no prior works have explored modeling of educational content (e.g. video) and their correlation with the learner's behavior (e.g. clickstream) in this context. We bridge this gap by devising a method to learn representation for videos and the correlation between videos and clicks. The results indicate that modeling videos and their correlation with clicks brings statistically significant improvements in predicting dropout.


\end{abstract}

\section{Introduction}
\label{sec:intro}
One of the most important challenges in MOOCs is to identify students most at risk with the goal of providing supportive interventions. Considering the huge number of users taking courses in MOOCs, it is necessary to automate the at-risk identification process. While there are several data sources to leverage in MOOCs, such as forum posts, assignments, and clickstreams, a recent study \cite{gardner2018dropout} proves that clickstream-based features are significantly better in dropout prediction. However, the raw clickstream data are too fine-grained to build a meaningful representation for downstream classification models, e.g. a dropout prediction model.

Many researchers have offered techniques to transform the raw clickstream data into the structured feature representation acceptable to existing statistical or machine learning models. One way is to use hand-crafted features, e.g. the count of each click type \cite{halawa2014dropout,DBLP:journals/ce/LykourentzouGNML09,Yang2013TurnOT,nagrecha2017mooc,whitehill2015beyond}. However, some studies use raw clickstream data without this feature engineering, claiming that it removes an important sequential pattern of the clickstream \cite{fei2015temporal,whitehill2017delving,wang2017deep,kim2018gritnet}. Still, these end-to-end models may overlook important patterns in the data by taking only a single objective into account. This naturally leads researchers to consider unsupervised methods to capture meaningful patterns under multiple objectives \cite{SinhaJLD14,Sinha2014}.

On top of this line of work, we hypothesize that the modeling of educational content (e.g. video) and their correlation with the learner's behavior (e.g. clickstream) should help dropout prediction. For instance, let's assume we observe a sequence of clicks indicating the struggles of learners (e.g. stopping the video). However, the same click pattern should be interpreted in a different way depending on the difficulty of educational contents. Nevertheless, no studies have been done in this direction. Thus, to test our hypothesis, we present dropout prediction model to explicitly capture the correlation between clicks and videos. We also propose a method to learn representation for clicks and videos in an unsupervised fashion. Our experimental results prove the benefit of modeling the correlation between clicks and videos and pretrained embeddings for click n-grams.

\section{Related Work}
\label{sec:related_work}


A substantial number of studies have predicted how likely a user is to drop out on a variety of MOOC datasets. However, many studies are based upon easily interpretable and static features such as grade \cite{halawa2014dropout,DBLP:journals/ce/LykourentzouGNML09,whitehill2015beyond}, demographics \cite{DBLP:journals/ce/LykourentzouGNML09}, forum posts \cite{Yang2013TurnOT}, etc \cite{Yang2013TurnOT}. Nevertheless, the majority of user interaction data in MOOCs is in the form of hardly interpretable clickstreams, and clickstream-based features are recently proved to be superior to other features in dropout prediction \cite{gardner2018dropout}. Our work aims to utilize rich and valuable information on user states from clickstreams, which could explain the chance of dropout.

A few works have still made use of clickstreams to predict dropout. One of main challenges for using clickstream data is to convert clickstream data into fixed-length and meaningful representation for downstream classification models. The prevalent solution to this challenge is to use hand-crafted features extracted from clickstream \cite{li2016dropout,amnueypornsakul2014predicting,taylor2014likely,kloft2014predicting}. These features include forum-related variables (e.g. the number of forum views), assignment-related variables (e.g. the number of submissions), and activity-related variables (e.g. the number of page/video views) \cite{nagrecha2017mooc,whitehill2015beyond}. However, these features inevitably lose temporal patterns in clickstreams and introduce unintentional biases due to the researchers' subjectivity. On top of that, some of the hand-engineered features are not platform-agnostic, which requires us to modify the feature extraction methods depending on the platform.


To resolve these issues, some researchers have attempted to predict dropout without manual feature engineering processes. They adopt deep neural network models since these are proved to be surpassing feature extractors regardless of domains even if they take the data in its raw form as an input. Given the sequential nature of clickstream, sequence models, such as recurrent neural network (RNN) and long short-term memory (LSTM), are the most popular choices \cite{fei2015temporal,whitehill2017delving}, while some researchers use more sophisticated deep neural network models \cite{wang2017deep,kim2018gritnet}. While these end-to-end models take the human out of the loop, they often lose the important signals from the data by optimizing only a single objective \cite{glasmachers2017limits}. In contrast, our representation learning framework complements end-to-end models by building meaningful representations for click sequence in an unsupervised manner while preserving temporal information in clicks.



The most similar works to ours in this regard focus on constructing cognitively meaningful representations from the clickstream. One approach is to build a combined representation of clickstream and discussion forum footprints using a set of graph metrics \cite{Sinha2014}. Another recent work groups raw click sequences into predefined behavioral categories and measures the degree of information processing as a proxy for the concentration \cite{SinhaJLD14}. However, unlike our method, none of them explicitly models the correlation between clickstream and video. Moreover, both of them rely on the hand-wavy design of behavioral category or taxonomy that provides us with the interpretability at the cost of missing important temporal signals. 

\section{Proposed Method}
\label{sec:proposed_method}
In this section, we describe our proposed network for dropout prediction capturing the correlation between learning contents (video) and learner's behavior (click sequence). Then, we present two approaches for learning click n-grams and video representation in an unsupervised fashion.

\begin{figure}[!htb]
	\centering
	\includegraphics[width=.95\linewidth]{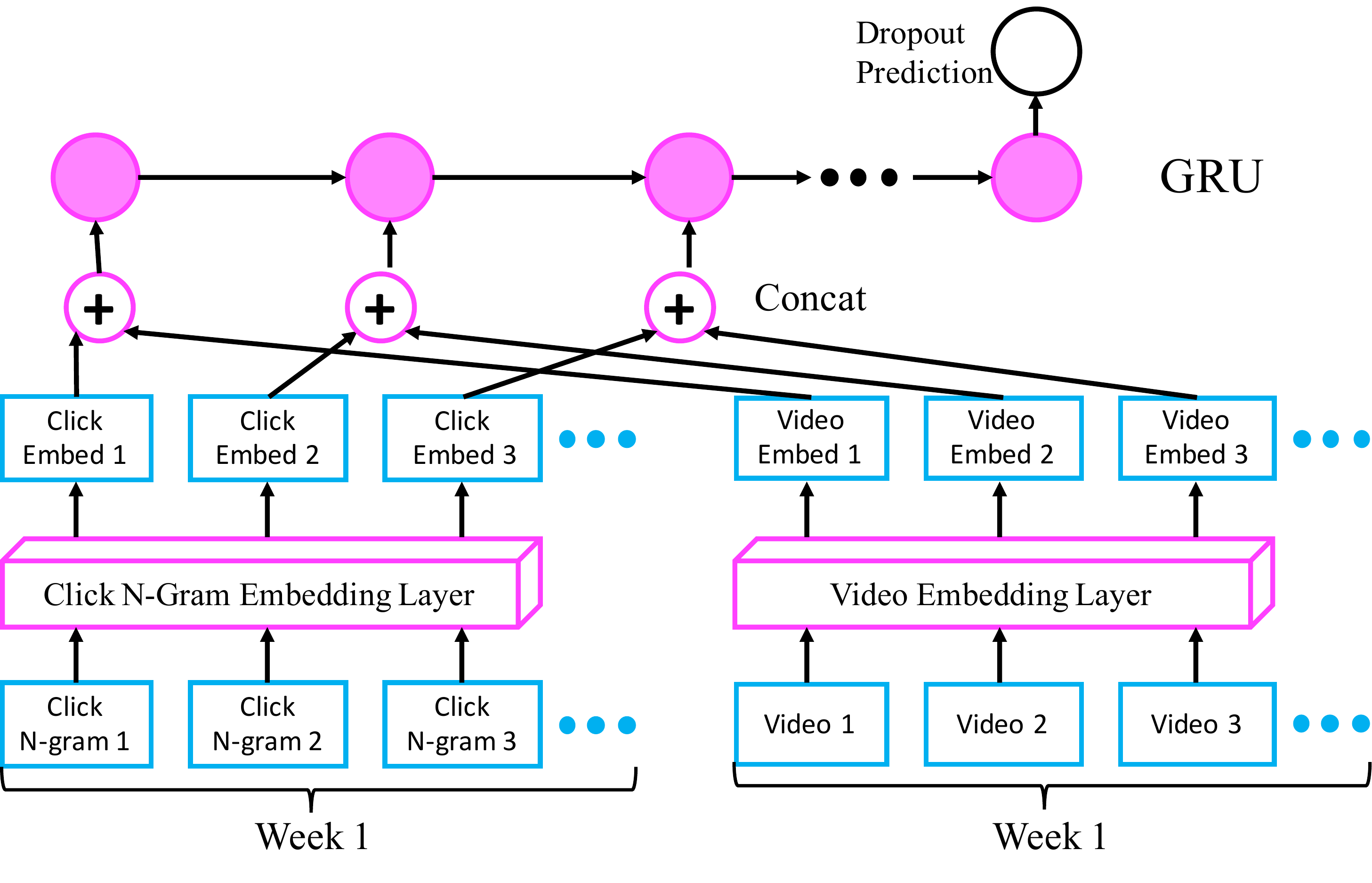}
	\caption{Dropout prediction network.}
	\label{fig:network}
\end{figure}

\subsection{Dropout Prediction Network}
\label{sec:method:baseline}

Given a clickstream data, we model how a learner's state evolves week by week
until he completes the course, or drops out of it.
Our intuition for the proposed network is that 
whether a student drops out or not is strongly associated with 
(1) his learning experiences or activities, and (2) the characteristics of the learning material.
For instance, a student's current understanding of the course topics,
how effective a student's learning strategy has been, 
how difficult the learning material is, and how well the content is presented
could all play a role in the decision of dropping out of a course.

In order to model a student's learning activity, our model considers
the sequential relationship among clicks. 
More specifically, our model uses a click $ n $-gram, instead of a single click,
as a unit for modeling a user activity based on the intuition that
a single click is far too fine-grained, and a user activity could be better represented by a click sequence.
We define a click $ n $-gram to be a sequence of 
$ n $ consecutive clicks where each click $ c $ corresponds to 
one of ten predefined user clicks (see \Cref{sec:exp:dataset} for details).
%
%
Our model also considers the characteristics of 
the learning material by receiving video information through a dedicated layer.
Further, we model how the two closely-related factors, which we capture 
by way of click $ n $-grams and videos, affect the student's
learning experience over time by combining them.


\Cref{fig:network} depicts our proposed network.
Given a click stream for the current week, we generate a sequence of 
click $ n $-grams and the corresponding video ids, and transform them into
click n-gram and video representations via two separate embedding layers.
Click n-gram and video embeddings are then concatenated to be fed into
a gated recurrent unit (GRU) layer, which learns the transition of a student state
for the current week, and outputs whether a student will drop out within the next week. 
Note that two embedding layers can be initialized with the weights pretrained by our unsupervised pretraining methods that follows.

To train our model, we minimize the margin ranking loss due to the imbalanced labels. We pair one positive instance (dropout) of the user with other negative instances of the same user. The objective function is 
\begin{align}
\min_{\bW} \frac{1}{T} \sum_{t=1}^{T} \max{(0, -(P_{pos} - P_{neg}) + M)}
\end{align}
where $\bW$ is composed of all the model parameters; $T$ is the number of pairs; $P_{pos}$ and $P_{neg}$ are the probability of dropout for positive and negative instance computed from our network; and $M$ is margin, which is set to 0.5.

\begin{figure*}[!ht]
	\begin{subfigure}[t]{0.40\textwidth}
		\includegraphics[width=.98\linewidth]{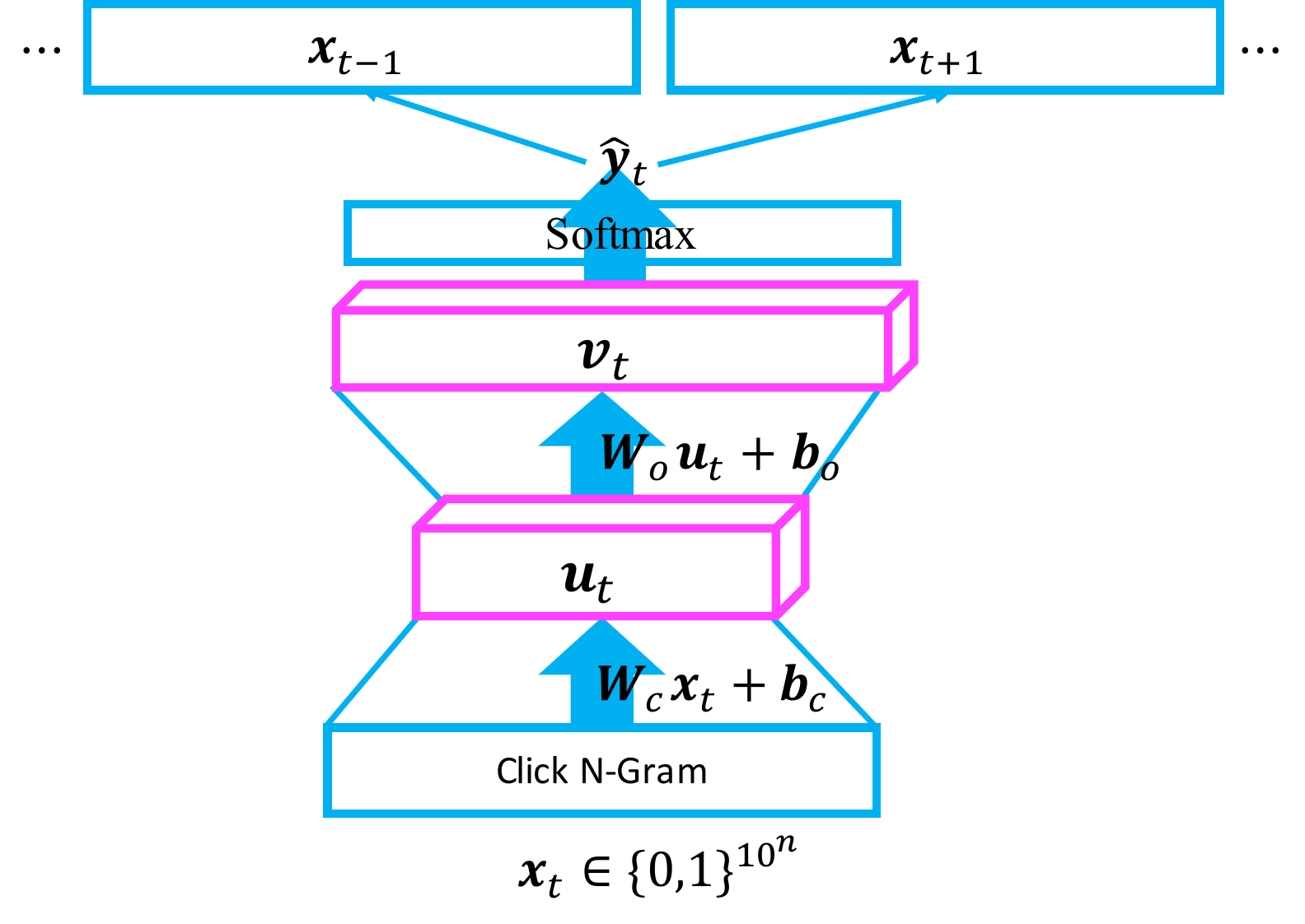}
		\caption{Pretraining click n-gram embeddings.}
		\label{fig:clickngram_pretraining}
	\end{subfigure}
	\begin{subfigure}[t]{0.60\textwidth}
		\includegraphics[width=.98\linewidth]{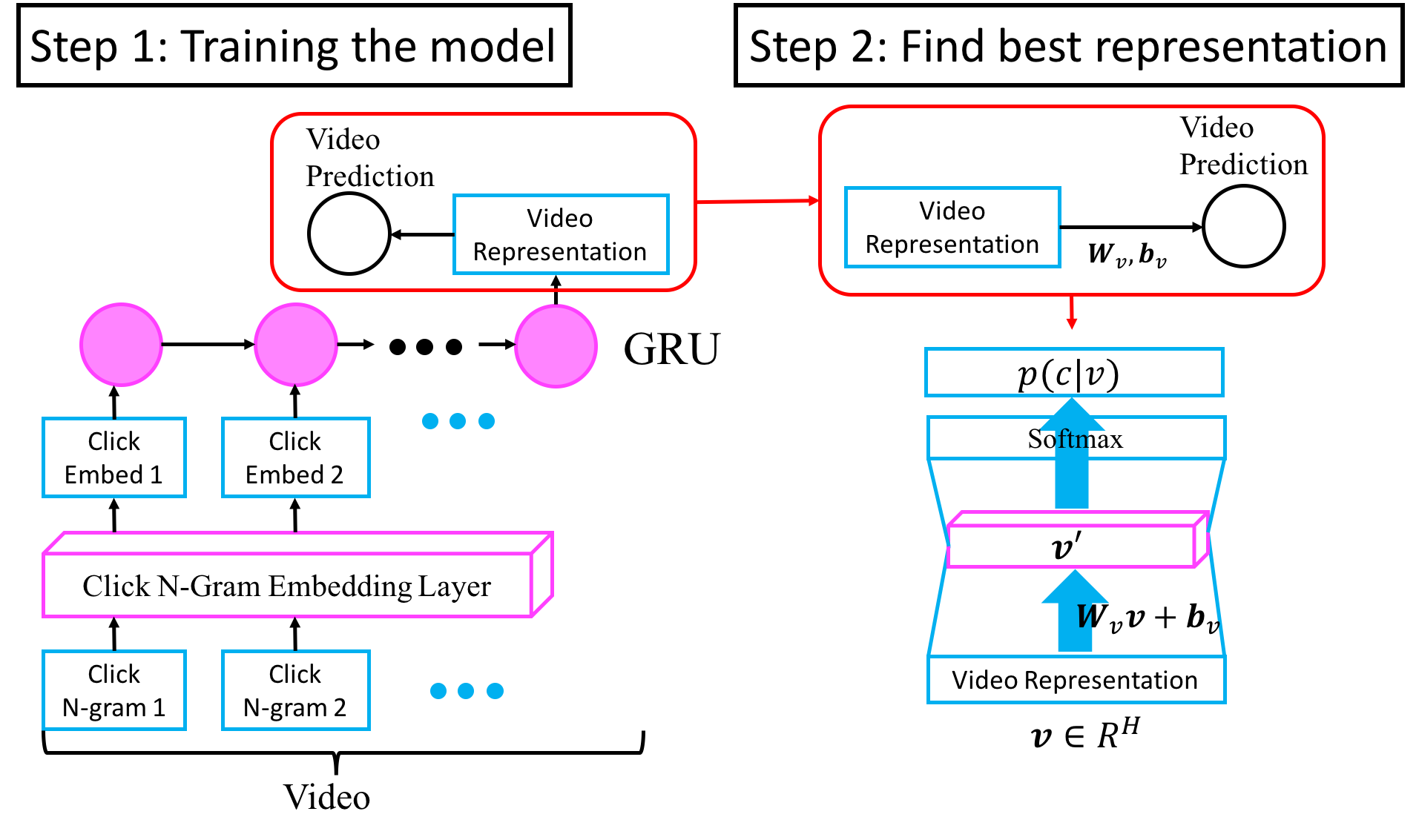}
		\caption{Pretraining video embeddings.}
		\label{fig:video_pretraining}
	\end{subfigure}
	\caption{Pretraining embeddings of click n-grams and videos.}
\end{figure*}

\subsection{Pretraining Click N-Gram Embeddings}
\label{sec:method:clickngram}

Given a sequence of clicks, we aim to capture the latent relationships between learners' consecutive clicks
via learning a representation of the clickstream that reflects the inter-click sequential information.
We train a multi-layer perceptron (MLP) using the following intuition:
Learning is a continuous process for each student. Thus a representation corresponding to a student's activity
at some point should be able to predict the student's activities in both the recent past and the near future.

Based on this intuition, we train the MLP architecture shown in \Cref{fig:clickngram_pretraining}.
A click $ n $-gram is represented as a one-hot vector of length $ 10^n $.
Given a click $ n $-gram $ \mathbf{x}_t \in \{0, 1\}^{10^n} $ for week $ t $, 
the first layer transforms it into a latent representation $ \mathbf{u}_t \in \mathbb{R}^{m} $ as follows:
\begin{align*}
\bu_t = \bW_c \bx_t + \bb_c
\end{align*}
where $\bW_c \in \R^{m \times 10^n}$ is the weight matrix for encoding a click $ n $-gram, and $\bb_c \in \R^{m}$ is the bias vector.
Then the second layer constructs a vector $\bv_t \in \R^{10^n}$ which is of the same size as the input $ \mathbf{x}_t $:
\begin{align*}
\bv_t = \bW_o \bu_t + \bb_o
\end{align*}
where $\bW_o \in \R^{10^n \times m}$ is the weight matrix for decoding an encoded click $ n $-gram, and $\bb_o \in \R^{10^n}$ is the bias vector.
Then, given the reconstructed representation $ \bv_t $, we minimize a cross entropy error as follows:

\begin{align*}
&\min_{\substack{\bW_{c},\bW_{o}, \\ \bb_{c},\bb_{o}}}\!\frac{1}{T}\!\sum_{t=1}^{T} \sum_{\substack{-w \le i \le w \\ (i \ne 0)}}\!-\bx_{t+i}^\top\!\log \hat{\by}_t\!-\!\left(\mathbf{1}\!-\!\bx_{t+i} \right)^\top\!\log(\mathbf{1}\!-\!\hat{\by}_t)\\
&\text{where~} \hat{\by}_t = \frac{\exp\left( \bv_t \right)}{ \sum_{j=1}^{10^n} \exp\left( \bW_o[j,:] \bu_t + \bb_o[j] \right)}, 
\text{~$ w $ is context}&
\end{align*}
window size, $T$ is the total number of weeks, $ \mathbf{1} $ is an all-one vector, and $ \exp $ is the element-wise exponential function.

\subsection{Pretraining Video Embeddings}
\label{sec:method:video}

In order to pretrain video embeddings, we have two steps: training the pretraining model and finding the best representation vector for each video. For the first step, we build up and train the model to predict the video label given consecutive click n-gram sequence for that video as in \Cref{fig:video_pretraining}. We use pretrained click embeddings and GRU to construct final video representation, which is the last hidden state of GRU. Based on this final video representation, our model makes a prediction on video label for given click n-gram sequence.

In the second step, our goal is to find the video representation $v$, which maximizes the probability of each video label given trained weight and bias of the last layer of this model, $W_v$ and $b_v$. Let $v_{max}^{(i)}$ be the best video representation for video $i$ and $c$ be the true video label, then it is expressed as follows:
\begin{align*}
&v_{max}^{(i)} = \mathop{\mathrm{argmax}}_{v}\ {p(c = i|v)}\\
&\text{where~} p(c = i|v) = \mathop{\mathrm{softmax}}(W_vv + b_v)_i
\end{align*}
We can easily get $v_{max}^{(i)}$ by the family of gradient descent algorithms and use it as the embedding vector of video $i$ in our dropout prediction network.
%


\section{Experiment}
\label{sec:experiment}
\subsection{Dataset}\label{sec:exp:dataset}
The dataset we use is collected from the Coursera\footnote{\url{https://www.coursera.org/}}, the top-ranked MOOC platform with more than 28 million users and 2,000 online courses, 
and is also used in \cite{DBLP:conf/lats/YangWHKR15}. The dataset includes 
clickstream data that contain clicks of Coursera learners who took video lectures of the Microeconomics course for maximum 12 weeks.
It includes 2,709,053 clicks collected from 48,090 users. 
Clicks are divided into 10 categories: 
\textit{Pageview}, \textit{Quiz}, \textit{Forum}, \textit{Play}, \textit{Pause}, \textit{SeekFwd}, \textit{SeekBwd}, \textit{RateFaster}, \textit{RateSlower}, and \textit{Stalled}.

\begin{table}[t!]
	\centering
	\caption{{\bf Dropout prediction performance of models that use different types of information and pretrainined embeddings.}}
	\label{table:performance}
	\begin{adjustbox}{max width=\textwidth}
		\begin{threeparttable}
			\begin{tabular}{l|c}
				\toprule
				\multicolumn{1}{c|}{Model} & AUC \\
				\midrule
				Click 4-gram & ~~0.740~~\\
				Click 4-gram (pretrained) & ~~0.757~~\\
				Click 4-gram (pretrained) + Video & ~~0.784~~\\
				Click 4-gram (pretrained) + Video (pretrained) & ~~0.783~~\\
				\bottomrule
			\end{tabular}
		\end{threeparttable}
	\end{adjustbox}
\end{table}

\subsection{Results}\label{sec:exp:result}

We evaluate the performance of our proposed method by measuring AUC on the dropout prediction task.  Each of four models listed in \Cref{table:performance} uses different types of information and pretrained embeddings. The first model, Click 4-gram, does not use the video embeddings, but the click 4-gram embeddings in \Cref{fig:network}. This is our state-of-the-art baseline for dropout prediction over weeks \cite{gardner2018dropout}. The second model, Click 4-gram (pretrained), also only uses the click 4-gram embeddings as the first model, but the embeddings are initialized with the pretrained embeddings shown in \Cref{fig:clickngram_pretraining}. The third model, Click 4-gram (pretrained) + Video, uses both the click 4-gram and video embeddings while only the click 4-gram embeddings are initialized with the pretrained embeddings in \Cref{fig:clickngram_pretraining}. The last model, Click 4-gram (pretrained) + Video (pretrained), uses both embeddings that are initialized with the pretrained embeddings in \Cref{fig:clickngram_pretraining} and \Cref{fig:video_pretraining}.

We see a statistically significant increase in AUC ($p < 0.05$) from the first to the second model, which indicates pretraining click n-gram embeddings captures useful temporal relationships between click n-grams for dropout prediction. In addition, another statistically significant increase in AUC ($p < 0.05$) from the second to the third model tells that video embeddings capture meaningful correlation between videos and clicks. However, there is no improvement from the third to the fourth model, which concludes pretraining video embeddings does not help dropout prediction. We conjecture that pretraining video embedding barely learns the expressive video representation because it is too difficult to predict a video label from the clickstream. Or clickstream could be too noisy to provide meaningful information for predicting a video label. We leave designing a better objective for pretraining video representation as our future work.

\bibliographystyle{aaai}
\bibliography{bsjeon}

\end{document}